\documentclass{article}
\usepackage{spconf,amsmath,graphicx}
\usepackage{threeparttable}
\usepackage{multirow}
\usepackage{booktabs}

\usepackage{color}

\title{Automated Segmentation of Pulmonary Lobes using Coordination-Guided Deep Neural Networks}
\name{Wenjia Wang$^{1}$, Junxuan Chen$^{3}$, Jie Zhao$^{2}$, Ying Chi$^{3}$, Xuansong Xie$^{3}$, Li Zhang$^{*, 1, 2}$\thanks{\hspace{-0.4cm}$^{\star}$Corresponding authors: Li Zhang, zhangli\underline{ }pku@pku.edu.cn;}, Xiansheng Hua$^{*,3}$\thanks{\hspace{2.5cm}Xiansheng Hua, xiansheng.hxs@alibaba-inc.com.}}
\address{$^{1}$ Center for Data Science, Peking University, Beijing, China;\\
$^{2}$ Beijing Institute of Big Data Research, Beijing, China;
$^{3}$ Alibaba Group, Hangzhou, China}
\begin{document}
%
\maketitle
\begin{abstract}
The identification of pulmonary lobes is of great importance in disease diagnosis and treatment. A few lung diseases have regional disorders at lobar level. Thus, an accurate segmentation of pulmonary lobes is necessary. In this work, we propose an automated segmentation of pulmonary lobes using coordination-guided deep neural networks from chest CT images. We first employ an automated lung segmentation to extract the lung area from CT image, then exploit volumetric convolutional neural network (V-net) for segmenting the pulmonary lobes. To reduce the misclassification of different lobes, we therefore adopt coordination-guided convolutional layers (CoordConvs) that generate additional feature maps of the positional information of pulmonary lobes. The proposed model is trained and evaluated on a few publicly available datasets and has achieved the state-of-the-art accuracy with a mean Dice coefficient index of 0.947 $\pm$ 0.044.
\end{abstract}
\section{Introduction}
\label{sec:intro}

Human lung is divided into five pulmonary lobes, which are served by independent bronchial and vessel trees. Many diseases are associated with specific lobes. Measuring a pathological lung at lobar level is therefore of great help to diagnose and assess different lung diseases. An accurate segmentation of pulmonary lobes is thus necessary. However, there are several challenges to the lobe segmentation. First, lobar boundaries defined by pulmonary fissures are often partially invisible from the CT scans. Furthermore, severe shape disorders of specific lobes may occur during pathological progress of the lung diseases. These problems has greatly limited the development of automated segmentation of pulmonary lobes.

Previously, several research groups have attempted to perform pulmonary lobe segmentation. A few unsupervised methods are reported, including watershed transformation \cite{Lassen2013Automatic}, graph-cuts \cite{giuliani2018pulmonary}, b-splines\cite{doel2012pulmonary}, surface fitting \cite{Bragman2017Pulmonary} and semi-automated segmentation framework \cite{lassen2011lung}. These methods use anatomical information as prior knowledge, including the segmentation of airways, vessels and fissures, and then generate final segmentation of pulmonary lobes. However, as described above, segmentation of airways, vessels and fissures are not always reliable. 

As encouraged by its recent success across the computer vision tasks, deep learning based methods are reported to segment pulmonary lobes. George et al. \cite{george2017pathological} propose a method that couples deep learning with the random walker algorithm. They employ the progressive holistically-nested network (P-HNN) model to identify potential lobar boundaries, and then delineate lobar boundaries using a random walker algorithm. Ferreira et al. \cite{ferreiraend} train an end-to-end deep learning model, known as Fully Regularized V-Net (FRV-Net), to segment five pulmonary lobes. However, in the absence of global/positional information of lobes, these method often generate incorrect segmentation, such as misclassification of different lobes and false response outside of lung area.

We propose an automated segmentation of pulmonary lobes using coordination-guided deep neural networks from chest CT scans. It is a fully end-to-end 3D deep learning approach without heavy post-processing schemes. In order to improve the accuracy of the proposed segmentation, we exploit lung segmentation as pre-processing. To further reduce the misclassification of different pulmonary lobes, we adopt coordination-guided convolutional layers (CoordConvs) which contain positional information of different pulmonary lobes. We evaluate the performance of the proposed method on 4 different data sets by several measuring metrics. The experiments show the superior performance of the proposed method compared to previous state-of-the-art methods.

\section{Method}
\label{sec:format}

\begin{figure}[htb]
\centering
\includegraphics[width=1\linewidth]{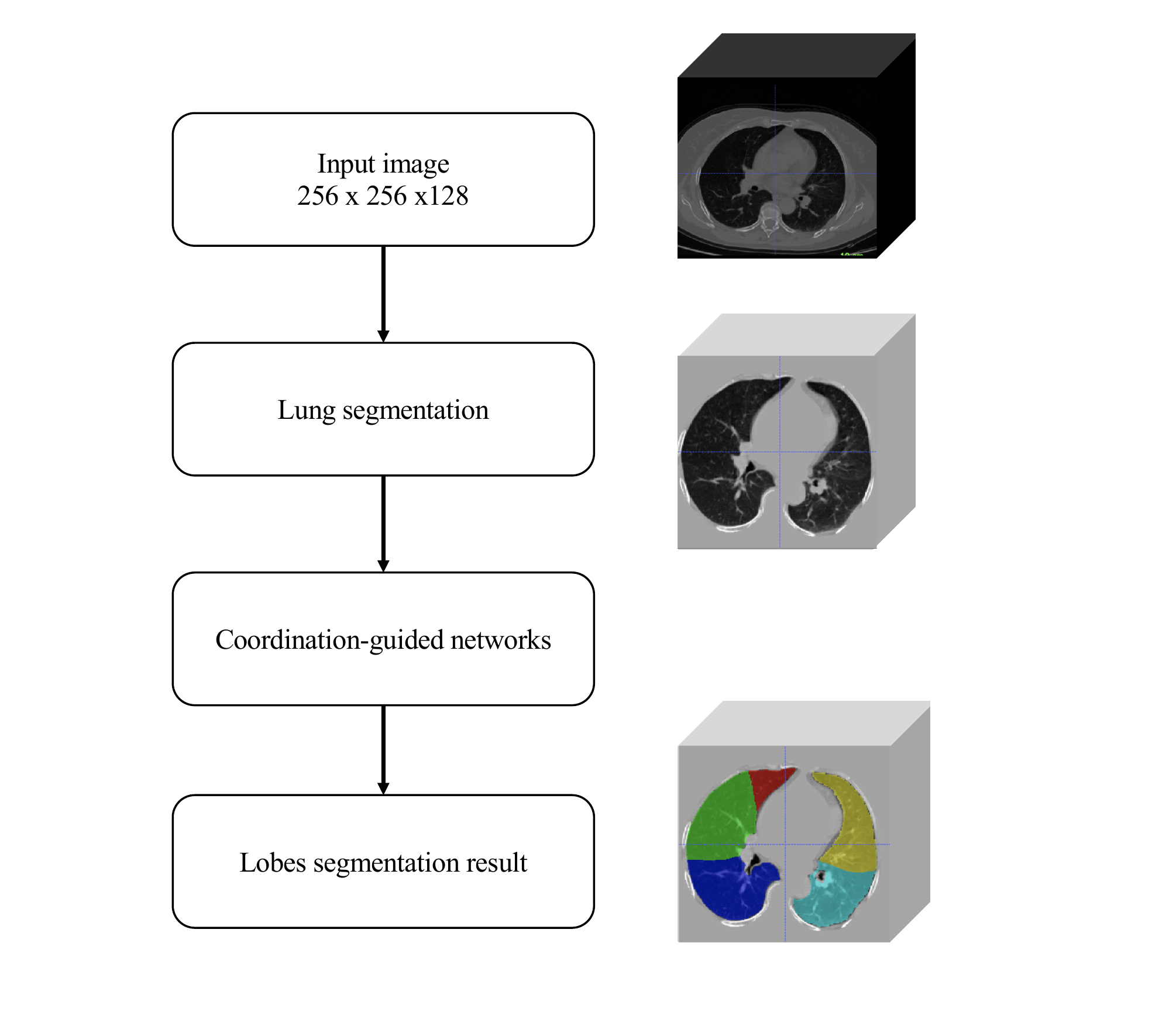}
\caption{The flowchart of the proposed method}
\label{flowchart}
\end{figure}

The architecture of this method is based on the  volumetric convolutional neural network (V-net)\cite{Milletari2016V}, which is widely used on 3D biomedical image segmentation. The input images are first downsampled to a size of 256$\times$256$\times$128 and a 2D automated lung segmentation is then applied to extract the lung area. After that, we add CoordConv layers to the decoding path of V-net. The proposed model produces a voxel-wise prediction for the five target lobar classes.  Figure \ref{flowchart} shows the flow chart of this method.

\subsection{V-net}
\begin{figure}[htb]
\centering
\includegraphics[width=1\linewidth]{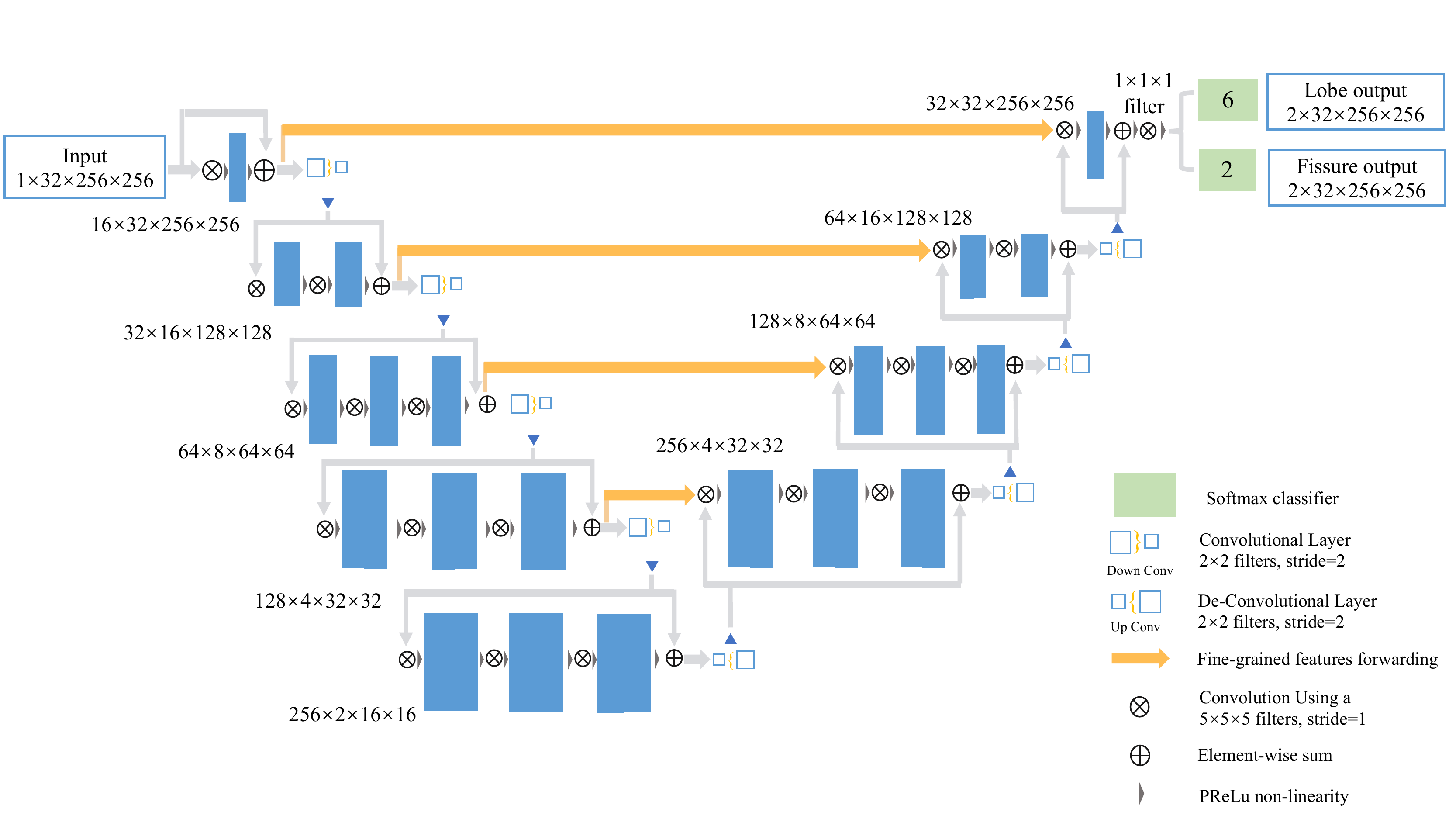}
\caption{vnet architecture}
\label{vnet architecture}
\end{figure}

V-net is widely used in 3D biomedical image segmentation tasks and achieves relatively good performance. As shown in Fig \ref{vnet architecture}, the left part of the network consists of a compression path, while the right part recovers the signal with its original size. Appropriate padding strategies are applied to maintain the shape of feature maps. The V-net uses skip connections concatenating feature maps to recover image details between the encoding and the decoding paths. To the contrary of original settings, we choose Parametric Rectified Linear Unit (PReLU) as the non-linear activation than Rectified Linear Unit (ReLU) to alleviate vanishing gradient problems. The last layer is a convolutional layer with kernel size of 1 and followed by a soft-max activation function which generates the probability maps of 5 lobe classes and 1 background class in a one-hot fashion.
\subsection{CoordConv Layers}
\begin{figure}[htb]
\centering
\includegraphics[width=1\linewidth]{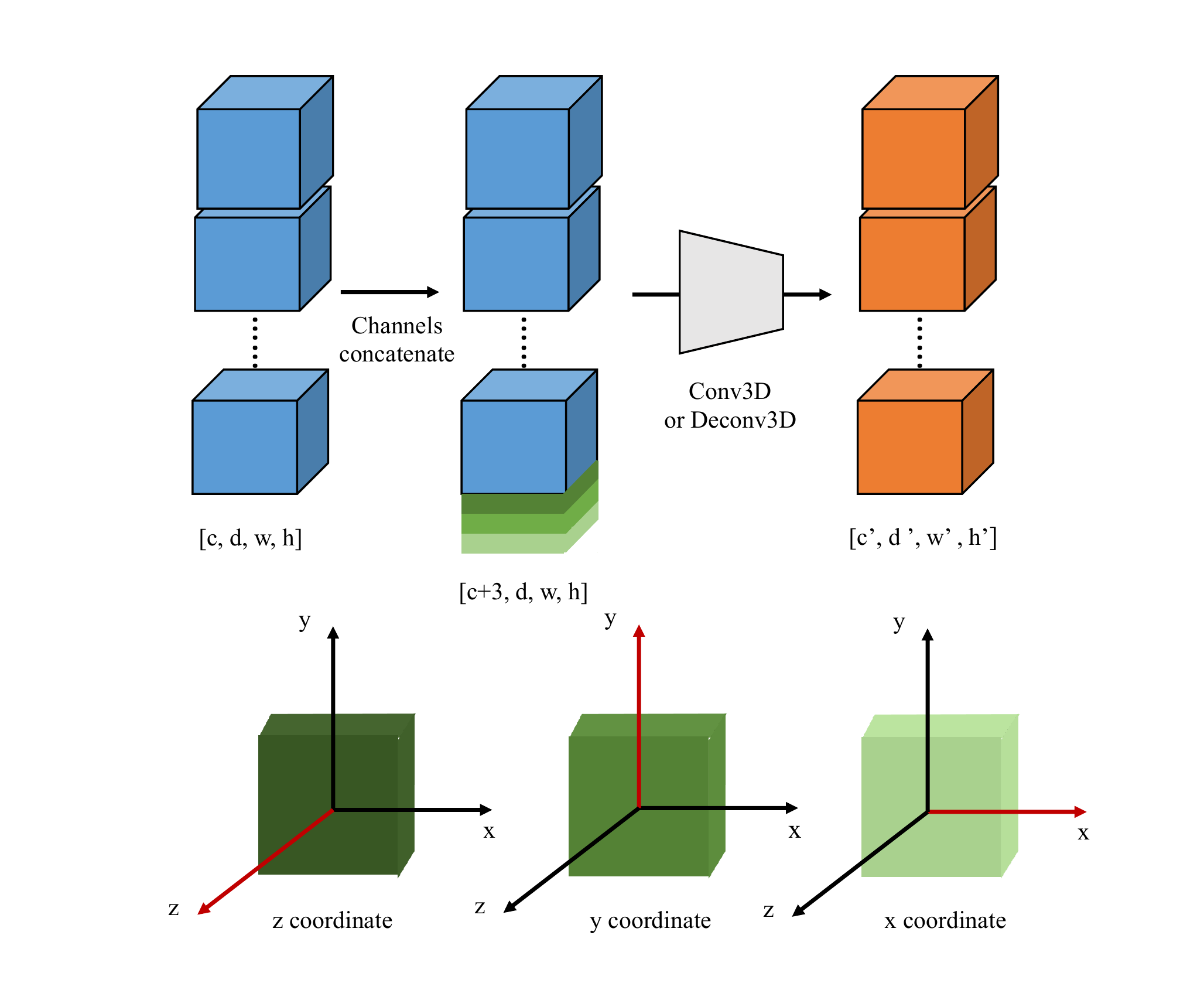}
\caption{The schematic diagram of 3D CoordConv layer}
\label{coordmap}
\end{figure}

The right lung normally comprises three lobes (upper, middle, lower) and the left lung normally has two lobes (upper and lower).  Although there are individual differences in chest CT scans, the lobes are distributed with distinct positional features. Misclassification often occurs if positional information of different lobes are not taken into account. However, information transfer between classic convolutional layers is usually limited within the receptive field of the layers, which is restricting the capability of classic convolutional layers to represent a global/positional information. In this work, we adopt an novel structure, a coordination-guided convolutional layer (CoordConv layer), to address this issue\cite{Liu2018An}. The CoordConv is a simple extension to the classic convolutional layer, integrating positional information by adding extra coordinate channels. As shown in Figure \ref{coordmap}, 3 extra channels are added respectively to represent x, y, and z coordinates of the input 3D images. We add CoordConv layers in the last transition in the decoding path and the values of coordination channels are normalized to the range from -1 to 1.

\subsection{Model Regularization and Loss Design}
In order to avoid the problem of overfitting we have implemented different regularizing techniques in our proposed model. First, dropout\cite{Srivastava2014Dropout} layers with a probability of 0.5 are applied while training and leads to a reduction of the sensitivity of specific neurons. Furthermore, we introduce data normalization during training process, such as batch normalization (BN)\cite{Ioffe2015Batch}. However, due to the memory limitation, the batch size is usually small when training a 3D convolutional neural network, resulting in inaccurate statistics estimation and rapid growth of BN error. Thus, we use group normalization\cite{Wu2018Group} that divides the channels into groups and computes within each group the mean and variance for normalization. 

We use the negative value of Dice coefficient\cite{Sudre2017Generalised} (called Dice loss) as the criterion to optimize the model parameters. The Dice loss is defined as,
\begin{equation} \label{lobedice_loss}
\scalebox{0.8}
 D_{lobes} = \sum\limits_{c}D_{lobes}^c=\sum\limits_{c}(-\frac{2\sum\limits_{i}p^c(i)g^c(i)}{\sum\limits_{i}p^c(i)+\sum\limits_{i}g^c(i)+\gamma })
\end{equation}
where $p^c(i)$ and $g^c(i)$ represent predicted value and groundtruth value at position $i$, respectively; $i$ iterates the entire image; $c$ denotes the class number ($c=1,2,3,4,5$; 1-5 for different classes of pulmonary lobes. we don't consider the background). And $\gamma$ is 1e-5 to avoid zero-division.

To further enhance the performance of the proposed model, we consider the detection of lobar boundaries as an auxiliary task. We use the same lobes segmentation networks and add additional boundary loss to improve the ability of the model to distinguish the lobes. The boundary loss is defined as,
\begin{equation} \label{fissuredice_loss}
D_{boundary}=-\frac{2\sum\limits_{i}p^b(i)g^b(i)}{\sum\limits_{i}p^b(i)+\sum\limits_{i}g^b(i)+\gamma }
\end{equation}
Finally, the total loss of our proposed model is 
\begin{equation} \label{total_loss}
D_{total}=D_{lobes}+\lambda D_{boundary}
\end{equation}
In the experiment,  $\lambda$ is set to $1$.
\begin{table}[!t]
	\centering
	\fontsize{6.5}{10}\selectfont
	\begin{threeparttable}
		\caption{Comparison of different model structures in the proposed method.}
		\label{tab:performance_comparison}
		\begin{tabular}{lcccccc}
			\toprule[1pt]
			Techniques&right-up&right-mid&right-low&left-up&left-low&average\cr
			\midrule[1pt]
		    baseline &0.801&0.656&0.829&0.859&0.830&0.795\cr
		    &(0.147)&(0.208)&(0.107)&(0.092)&(0.128)&(0.171)\cr
		    \midrule
		    + coordmap&0.897&0.846&0.941&0.947&0.947&0.916\cr
		    &(0.092)&(0.117)&(0.034)&(0.047)&(0.028)&(0.045)\cr
		    \midrule
		    + group norm&0.921&0.868&0.952&0.952&0.954&0.929\cr
		    &(0.096)&(0.137)&(0.028)&(0.049)&(0.028)&(0.050)\cr
		    \midrule
		    Proposed&0.934&0.919&0.953&0.958&0.973&0.947\cr
		    &(0.090)&(0.104)&(0.030)&(0.047)&(0.030)&(0.044)\cr
			\bottomrule[1pt]
		\end{tabular}
	\end{threeparttable}
\end{table}

\section{Experiments and results}
\label{sec:pagestyle}

\subsection{Data}
We have included 343 chest CT scans in this work, including 71 cases from the LUNA16 \footnote{https://luna16.grand-challenge.org/} dataset, 195 cases from LKDS \footnote{https://tianchi.aliyun.com/getStart} dataset, 62 cases from Meinian One-health Health-care Holdings, 15 from Lobe and Lung Analysis 2011 (LOLA11) competition \footnote{https://lola11.grand-challenge.org/} .  All CT scans are manually annotated by an experienced radiologist using 3D Slicer Platform (SlicerCIP). The value of slice spacing of these CT volumes varies from 0.5 to 1.5$mm$. Both healthy and pathological lungs are included.

\subsection{Experiments}
The model is implemented using Pytorch 0.4.0 package, and runs on NVIDIA Tesla P100 GPU with 16 GB of memory. We evaluate our results using five-fold cross validation. The performance evaluation metric is Dice coefficient index. In order to assess the effectiveness of the proposed techniques (CoordConvs, group normalization, and etc.), a few experiments are performed using different model settings with or without these techniques. All experiments are under the same set of hyper-parameters. We also compare our method with several previous state-of-the-art methods. 
\begin{table}[!t]
	\centering
	\fontsize{6.5}{10}\selectfont
	\begin{threeparttable}
		\caption{Performance comparison of our method with previous methods.}
		\label{other_comparison}
		\begin{tabular}{lcccccc}
			\toprule[1pt]
			Methods&right-up&right-mid&right-low&left-up&left-low&average\cr
			\midrule[1pt]
			Watershed[1]&0.920 &0.770 &0.910 &0.920 &0.890 &0.880\cr
			&(0.090)&(0.300)&(0.180)&(0.160)&(0.230)&( - )\cr
			\midrule
			PPLS + RM[6] &0.929 &0.824&0.887 &0.925 &0.945&0.888\cr
			&(0.057)&(0.142)&(0.239)&(0.244)&(0.175)&(0.116)\cr
			\midrule
			Alpha-Exp[2] &0.873 &0.714 &0.928&0.929&0.884&0.866\cr
			&(0.169)&(0.322)&(0.104)&(0.118)&(0.231)&( - )\cr
			\midrule
			FR-vnet[7]&0.931&0.869&0.941&0.948&0.941&0.926\cr
			&(0.070)&(0.110)&(0.077)&(0.065)&(0.070)&(0.055)\cr
			\midrule
			Proposed&0.934&0.919&0.953&0.958&0.973&0.947\cr
		    &(0.090)&(0.104)&(0.030)&(0.047)&(0.030)&(0.044)\cr
			\bottomrule[1pt]
		\end{tabular}
	\end{threeparttable}
\end{table}
\begin{figure*}[t]
\centering
\includegraphics[width=1\linewidth]{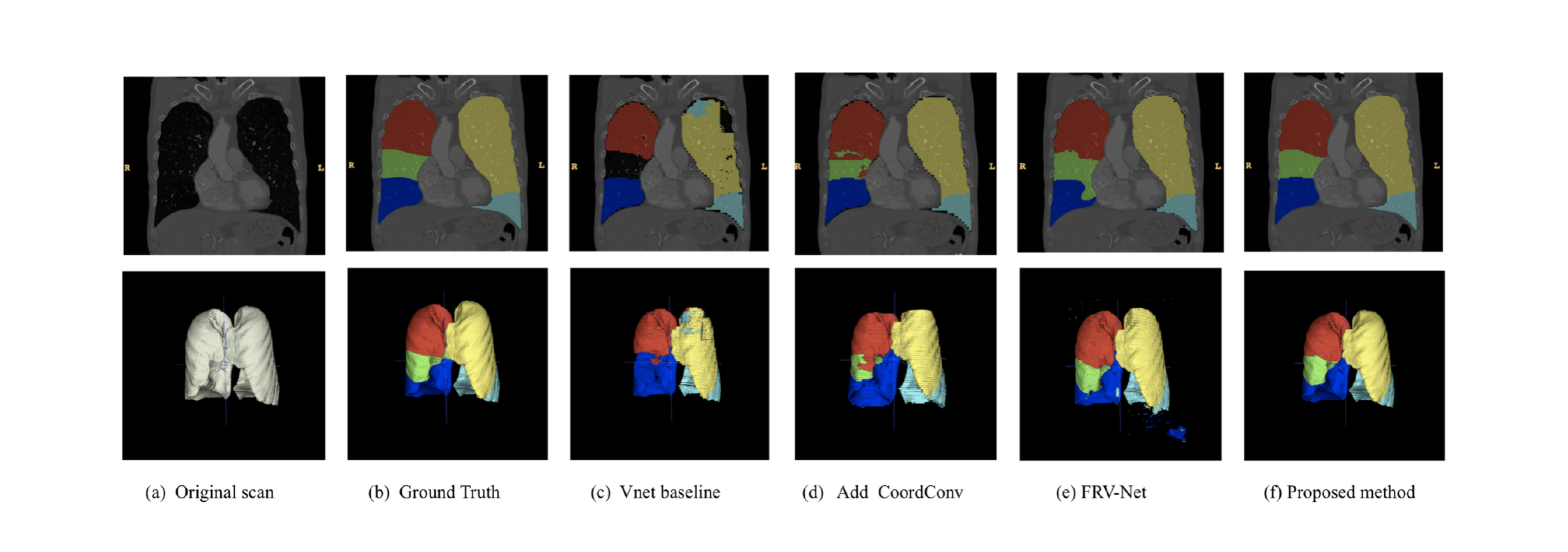}
\caption{Example of the segmentation results using different model settings}
\label{comparision figures}
\end{figure*}
\subsection{Results}
Table \ref{tab:performance_comparison} shows the quantitative performance of the proposed method in terms of mean $\pm$ std of Dice coefficient index. Per-lobe-Dice and overall mean Dice are showed in the Table. We also compare our proposed method with the state-of-the-art methods (see Table \ref{other_comparison}). Figure \ref{comparision figures} shows an example of our segmentation results in 2D and 3D. Previous methods suffer from two major issues to correctly identify the pulmonary lobes: 1) False segmentation outside of lung area (see Figure \ref{comparision figures}(e)), and 2) misclassification of different lobes (see Figure \ref{comparision figures}(c)). To address the first issue, a 2D automated lung segmentation(U-net\cite{Ronneberger2015U}) are performed as a preprocessing step to extract lung ROIs. This step not only effectively solve the problem of false segmentation outside of lung, but also improve the lobe segmentation due to the narrower searching space of the model. For the second issue, we add CoordConv layers to the standard convolutional networks. The positional information carried by CoordConv layers acts as ``soft constraint'' and guarantees that the segmented lobes are around correct location. Finally, the proposed model achieves an average per-lobe-Dice of 0.947, compared to the previously best state-of-the-art approach of 0.926. The overall processing time of our proposed model is 12s per case on average using one Nvidia Tesla P100 GPU. 

\section{Conclusion}
We propose an automated segmentation of pulmonary lobes using coordination-guided deep neural networks from chest CT scans. Several techniques are designed to improve the overall segmentation accuracy. First, an accurate lung segmentation is introduced to remove the false positive response outside of lung area. Second, CoordConv is added for a more effective transformation of the positional information. Finally, special designs of model regularization and loss function are performed. Experimental results demonstrate the superior performance of our proposed method. However, there are only thin slices in our cases. 

\label{sec:typestyle}

\section{Acknowledgment}
\label{sec:Acknowledgment}
This work is supported in part by the National Natural Science Foundation of China (NSFC) under Grants 81801778, 11831002 and the National Key Research and Development Program of China under Grant 2018YFC0910700.

\begin{table}[!t]
	\centering
	\fontsize{6.5}{10}\selectfont
	\begin{threeparttable}
		\caption{Performance comparison of our method with previous methods.}
		\label{other_comparison}
		\begin{tabular}{lcccccc}
			\toprule[1pt]
			Methods&right-up&right-mid&right-low&left-up&left-low&average\cr
			\midrule[1pt]
			FR-vnet[7]&0.931&0.869&0.941&0.948&0.941&0.926\cr
			&(0.070)&(0.110)&(0.077)&(0.065)&(0.070)&(0.055)\cr
			\midrule
			Proposed&0.934&0.919&0.953&0.958&0.973&0.947\cr
		    &(0.090)&(0.104)&(0.030)&(0.047)&(0.030)&(0.044)\cr
			\bottomrule[1pt]
		\end{tabular}
	\end{threeparttable}
\end{table}

\bibliographystyle{IEEEbib}
\bibliography{main}

\end{document}